\documentclass[aip,cha,twocolumn,10pt]{revtex4-1}

\usepackage{amsmath}
\usepackage{graphicx}
\usepackage{rotating}
\usepackage{multirow}
\usepackage{latexsym}
\usepackage{textcomp}
\usepackage{verbatim}
\usepackage{color}
\usepackage{pict2e}
\usepackage{bm}
\usepackage{subfigure}
\usepackage{everysel}
\usepackage{keyval}
\usepackage{ragged2e}
\usepackage{dsfont}
\usepackage{amssymb}
\usepackage{enumitem}
\usepackage{pstricks}
\usepackage{mathtools}
\usepackage{hyperref}

\usepackage{amsfonts}
\usepackage{algorithmic}
\usepackage{textcomp}
\usepackage{booktabs}
\usepackage{multirow}
\usepackage{caption}

\usepackage{algorithm}

\usepackage{graphicx}
\usepackage{subfigure}  % support sub-figure

\def\etal{{\it et al. }}

\begin{document}
% \history{Date of publication xxxx 00, 0000, date of current version xxxx 00, 0000.}
% \doi{10.1109/ACCESS.2018.DOI}

\title{Chaotic Genetic Algorithm and The Effects of Entropy in Performance Optimization}
\author{Guillermo Fuertes}
\affiliation{Facultad de Ingenier\'ia, Ciencia y Tecnolog\'ia, Universidad Bernardo O'Higgins, Santiago 8370993, Chile}
\affiliation{Facultad de Ingenier\'ia, Universidad de San Buenaventura, Ciudad de Cali, Colombia.}
\author{Manuel Vargas}
\affiliation{Facultad de Ingenier\'ia y Tecnolog\'ia, Universidad San Sebasti\'an, Bellavista 7, Santiago 8420524, Chile} 
\author{Miguel Alfaro}
\affiliation{Departamento de Ingenier\'ia Industrial, Universidad de Santiago de Chile, Santiago 9170124, Chile}
\author{Rodrigo Soto-Garrido}
\affiliation{Facultad de Ingenier\'ia y Tecnolog\'ia, Universidad San Sebasti\'an, Bellavista 7, Santiago 8420524, Chile} 
\author{Jorge Sabattin}
\affiliation{Facultad de Ingenier\'ia, Universidad Andres Bello, Antonio Varas 880, Santiago  7500971, Chile}
\author{Mar\'ia Alejandra Peralta}
\affiliation{Departamento de Ense\~nanza de las Ciencias B\'asicas, Universidad Cat\'olica del Norte, Coquimbo 1789002, Chile}

%\tfootnote{This paragraph of the first footnote will contain support 
%information, including sponsor and financial support acknowledgment. For 
%example, ``This work was supported in part by the U.S. Department of 
%Commerce under Grant BS123456.''}

% \markboth
% {Author \headeretal: Preparation of Papers for IEEE TRANSACTIONS and JOURNALS}
% {Author \headeretal: Preparation of Papers for IEEE TRANSACTIONS and JOURNALS}

%\corresp{Corresponding author: Manuel Vargas (e-mail: manuel.vargasg@usach.cl).}

\begin{abstract}
This work proposes a new edge about the Chaotic Genetic Algorithm (CGA) and the importance of the entropy in the initial population. Inspired by chaos theory the CGA uses chaotic maps to modify the stochastic parameters of Genetic Algorithm (GA). The algorithm modifies the parameters of the initial population using chaotic series and then analyzes the entropy of such population. This strategy exhibits the relationship between entropy and performance optimization in complex search spaces. Our study includes the optimization of nine benchmark functions using eight different chaotic maps for each of the benchmark functions. The numerical experiment demonstrates a direct relation between entropy and performance of the algorithm.
\end{abstract}


\maketitle

{\bf In the following paper, we used chaotic maps to improve the performance of the so called genetic algorithms, which have been widely used   in optimization and search problems. For instance, the genetic algorithms have been used to improve production processes and services in different areas of engineering, such as air transport, maritime transport, electric power distribution, logistics, electronic circuit design and production planning among other. The main result of this work was to established a closed relation between the entropy of the initial population and the performance of the modified genetic algorithm. Higher entropy of the initial population implies better solutions of the algorithm.

}

\section{Introduction}
\label{sec:introduction}
The evolutionary computation possesses a strong biological base, in its origins, the evolutionary algorithms consisted of generating a copy of the processes of natural selection, concept introduced by Charles Darwin.\cite{Fuertes2018}
Although the biological evolution has not been completely deciphered, there are some facts that have a strong experimental support, such as:

\begin{itemize}
	\item The evolution works more on the chromosomes (genotype), that on the organisms themselves (phenotype). Those chromosomes are considered as the organic tools that codify the life and the biological created form it is decoded from the information contained in the chromosomes.
	\item The natural selection is the mechanism that relates the genotype to the phenotype, giving a better life expectation and reproduction to the most adapted individuals.
	\item The evolution happens during the reproduction's stage.
\end{itemize}

There are three main points that due to the computational limitations in the origins of the GA were simplified and that can affect the process of evolution. We discussed over:

\begin{itemize}
	\item The populations are a dynamical system, therefore there is a constant fluctuation in the number of individuals, in contrast to the constant population used in the GA.
	\item The processes happen during the reproduction, as in the crossing and mutation. These processes are completely chaotic and do not depend on a single parameter.
	\item The characteristics of the initial population are not completely random. If the reproduction is chaotic the population will be chaotic as well.	
\end{itemize}

In Refs. \cite{Angelova2011,Karofotias} the authors studied the effect of constant size populations in the evolutionary computation. They asked the following question: ``does the status quo imply that the size of the population is not an important parameter?''. Inspired by nonlinear control mechanisms of the natural populations dynamical systems, they introduced algorithms with variable populations defined by chaotic maps, especially using the logistic function. This dynamical configuration increases the performance of the algorithm in getting better solutions.

Chaos theory studies dynamical systems that correspond to systems that evolve over time. The present state depends on the characteristics of the previous state, in contrast to random systems where it is impossible to predict the next state from the previous one.\cite{Fuertes2015a} 

Despite the advantages in using the GA, previous studies have shown a premature convergence into a local optimum.\cite{Pandey2014} Some researchers have introduced new techniques, by hybridizing the GA with others evolved algorithms.\cite{Wei2010a,Yanguang2010} These methods improve substantially the stability and adaptability of the algorithm to solve specific problems.\cite{Chatterjee2011,Song2010,Lu2013} Some recent researchers modify those hybridizations with the use of chaotic maps.\cite{Liang2015,Cui2017} 

Using chaotic maps to generate the initial population of the GA avoids the premature convergence to a local solution. These algorithms modified by chaotic series are the so called Chaotic Genetic Algorithms (CGA). The performances differ significantly between the GA modified by the use of chaotic maps CGA and GA using stochastic parameters. In Ref.\cite{Xue2018} Xue and collaborators incorporated chaotic maps in the artificial colonies' algorithm of bees and found better results than using the standard GA. 
Moreover, a CGA based on the Lorenz function was implemented in \cite{Ebrahimzadeh2013}, improving the performance of the Schaffer and Clonalg functions with respect to the usual GA.  In addition, there is a relation between the entropy of the populations and the results of the Swarm Intelligence model, indicating a direct relation between entropy and performance of the model.\cite{Liu2009} This fact motivated us to develop an approach for the optimization of non-convex functions, and the relation between the results obtained in the search of ideal global solution with the entropy of the initial populations generated by the chaotic maps.

The main goal of this work is to study the use of chaotic series for the generation of initial populations in the CGA. To analyze the performance of the modified algorithm we use nine benchmark non-convex continuous functions to measure the improvement compare to the traditional GA. In addition, the relation between the entropy of the initial population with the performance of algorithms is analyzed.

Using Lorenz chaotic function in the productive processes, Ebrahimzadeh and Jampour \cite{Ebrahimzadeh2013} obtained values of decisions during the crossing and mutation. Their results demonstrate an improvement in the performance of the GA with chaotic processes compare to the traditional genetic algorithm. In addition, they proposed other chaotic functions like the Henon, Logistic or Rossler function, that produced similar results. 

The work aforementioned \cite{Ebrahimzadeh2013} approaches from different edges the interaction between initial populations and its generation using chaotic maps. The researchers noticed an improvement in the obtained solutions where chaotic maps where used in the generation of the initial population. However, they were not conclusive in classifying the performance of every chaotic map and its relationship with the entropy.

% {\color{red} The conditions to maintain chaotic behavior and their unpredictability are guaranteed with the high maximum Lyapunov exponent and high entropy.\cite{Valtierra2017,ValtierraSanchezdelaVega2015,DelaFraga2014} These conditions allow to improve the modeling of continuous-time multi-scroll chaotic attractors.\cite{Tlelo-Cuautle2016} }

Entropy measures uncertainty in a series of information units .\cite{Bat-Erdene2017,Zhang2011a,Guo2014} Solteiro \etal \cite{SolteiroPires2013} remark the use of Shannon entropy in several studies, and the capability to provide useful information about the diversity of the population and the entropy's strategy to preserve diversity. The state of the algorithms is revealed by the distance among populations and can be measured by the Shannon entropy; a smaller entropy represents a saturated space and that the diversity be reduced.\cite{Lopez-Garcia2016}

Other related evolutionary algorithms enhanced by chaotic maps are described in \cite{Laboudi2012,Terki2016,Mojarrad2015}. In those algorithms, the generation of random values for the different parameters in the model is replaced by the application of chaotic maps for the generation of those parameters. Both, the quality of the solutions and the capacity to search in spaces removed from premature convergences, increase with this method.

Finally, recent studies referred to  genetic quantum algorithms compared with algorithms based on chaotic functions give complementary lines of research with preliminary similar results developed in this section.\cite{Teng2010}

\subsection{Complexity metrics}

A chaotic map is a dynamical system that produces chaotic variables. The chaotic behavior works as strangers attractors, which may have an enormous complexity, like Lorenz's attractor, which arises from the modeling of the climatic system.

\subsubsection{Lyapunov exponent}

These chaotic systems are characterized for having a strong dependence on their initial conditions. The  Lyapunov exponent shows, in an analytic way, this relation as follows:

Let us define $x(t)$ as an orbit that turns around the attractor and considered an initial condition $x(0)$ and a second initial condition to ${x}'(0)$, which is displaced of $x(0)$ by an infinitesimal distance $\delta x(0)$.

\begin{equation}
{x}'(0)=x(0)+\delta x(0)
\end{equation}
Then the Lyapunov exponent, $h$, is given by:

\begin{equation}
h=\lim_{t\rightarrow \infty}\frac{1}{t}\ln\left ( \frac{\delta x(t)}{\delta x(0)} \right )> 0
\end{equation}

For positive exponents, the trajectories of very close points get separated depending on the time in an exponential way. For our case, a positive Lyapunov exponent guarantees a chaotic behavior of the time series used in this study. 

\subsubsection{Shannon Entropy}
Shannon \cite{Shannon1948} introduced the concept of information entropy in 1948 to quantify the information contained in a message.\cite{Cheng2016a} The Shannon's entropy is defined from the probability distribution, where $p(x_i )$ denotes the probability of each state $i$. The parameter $K$ is a positive constant. This concept may be extended for two variables $\left ( x_i ,y_j \right )\in \left ( X,Y \right )$.

The measure of uncertainty of information \cite{Eguiraun2014} content in a system is calculated by the equation:

\begin{equation}
H\left ( X,Y \right )=-K\sum_{i,j} p\left ( x_i ,y_j \right ) \log p\left ( x_i ,y_j \right )
\end{equation}
% Related to the previous, Kolmogorov's entropy is defined as the sum of the Lyapunov exponents and represents the average speed of information loss in a temporary series. The entropy is a measure of the order; defined as the  number of possible configurations of the system.\cite{Teng2010}

\section{Materials and Methods}

\subsection{Description algorithm implemented}

The Chaotic Genetic Algorithm replaces random processes in the generation of the initial populations by chaotic processes. The numerical series built by chaotic maps, such as the Henon map, generate the initial population.\cite{Leontitsis2004} 
\begin{equation}
X_{n+1}=1-A*X_{n}^2+Y_{n}~~~~~;~~~Y_{n+1}=B*X_{n}
\end{equation}
where each $n$ corresponds to a given chromosome $(X_n,Y_n)$ of the initial population. In the case of flows (Lorenz and Rossler maps) we discretized the time to obtain the genes for the chromosomes. We used a time step of 0.01 and 0.1 for the Lorenz and Rossler maps respectively.  

We implement the CGA based on chaotic systems and test it on benchmarks functions such as the Beale and the Leon functions (see section \ref{sec:BF}).
\begin{align}
z_\text{(Beale)}\left ( x,y \right )=&\left ( 1.5 -x+xy\right )^{2}
+\left ( 2.25-x+xy^{2} \right )^{2}\notag\\
&+\left ( 2.625 -x+xy^{3} \right )^{2}    \\  
z_\text{(Leon)}\left ( x,y \right )=&100\left ( y-x^{2} \right )^{2}+\left ( 1-x \right )^{2}   
\end{align}

Others parameters of CGA, such as the population size, the probability of crossover, the probability of mutation and the number of generations are kept constant. 
\smallskip

\noindent Population Size = 100 \\
Probability Crossover = 0.8 \\
Probability Mutation = 0.03 \\
Generation = 50

\smallskip
The entropy $H$ is calculated for each initial population before starting the cycle of optimization with 50 generations. The populations are selected according to their fitness through the roulette wheel selection; better chromosomes have more chances to be selected.  
The criteria to terminate the simulation is by reaching the maximum number of iterations, then the algorithm success rate or performance $(P)$ has been used to compare the results. The performance is defined by \cite{Fuertes2018,Angelova2011}: 
\begin{equation}
P=100*\frac{N\left ( \textup{Optimal Results} \right )}{Nt\left ( \textup{Total Cycles} \right )}
\end{equation}
Others authors used average and standard deviation or the Wilcoxon statistical test to discuss the statistical results.\cite{Xue2018,Teng2010,Wang2010}

\subsection{Benchmark functions}
\label{sec:BF}

Siva and Radhika \cite{SivaSathya2013} defined a set of mathematical functions, the so called benchmark functions, to evaluate the performance of the evolutionary algorithms. Eiben and Smit \cite{Eiben2011a}
used these functions to evaluate the optimization in several evolutionary algorithm.

\begin{figure}[!t]
	\centering
	\hspace*{-0.55cm}
	\includegraphics[width=1.1\linewidth]{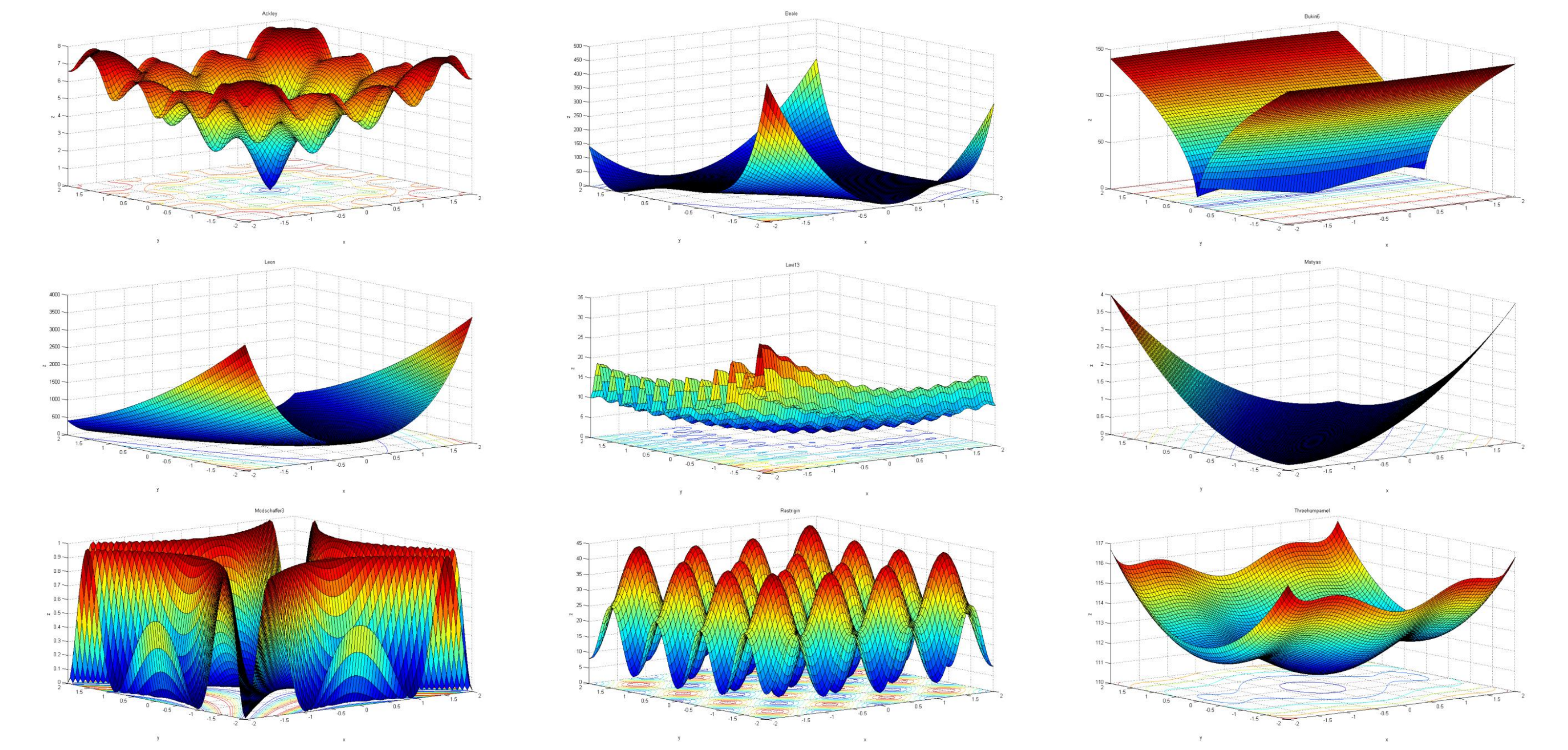}
	\caption{From left to right in your order the benchmark functions used in the present study: Ackley, Beale, Bukin6, Leon, Levi13, Matyas, Modschaffer2, Rastrigin and Treehupamel.}
	\label{fig:figura1}
\end{figure}

The functions are classified by: (1) separable or non-separable functions, (2) multimodal and (3) the dimensionality of the search space. Separable function with $n$ variables are rewritten as the sum of $n$ functions of just one variable. Multimodal functions have two or more local optima, if the local optima are randomly distributed in the search space the solution is more complex. Finally, more accuracy in search space signified differences in processing times. Figure \ref{fig:figura1} shows the benchmark functions used in the present study. These functions are: Ackley, Beale, Bukin6, Leon, Levi13, Matyas, Modschaffer2, Rastrigin and Treehupamel.

\subsection{Diagram of the algorithm proposed}
The GA uses random processes in the generation of the initial populations. The algorithms proposed in this study replace the stochastic processes for chaotic ones of deterministic character. The chaotic maps dynamical features, such us ergodicity, irreversibility and nonlinearity, allow to avoid premature convergences in the GA.

\begin{figure}[!t]
	\centering
	\includegraphics[width=\columnwidth]{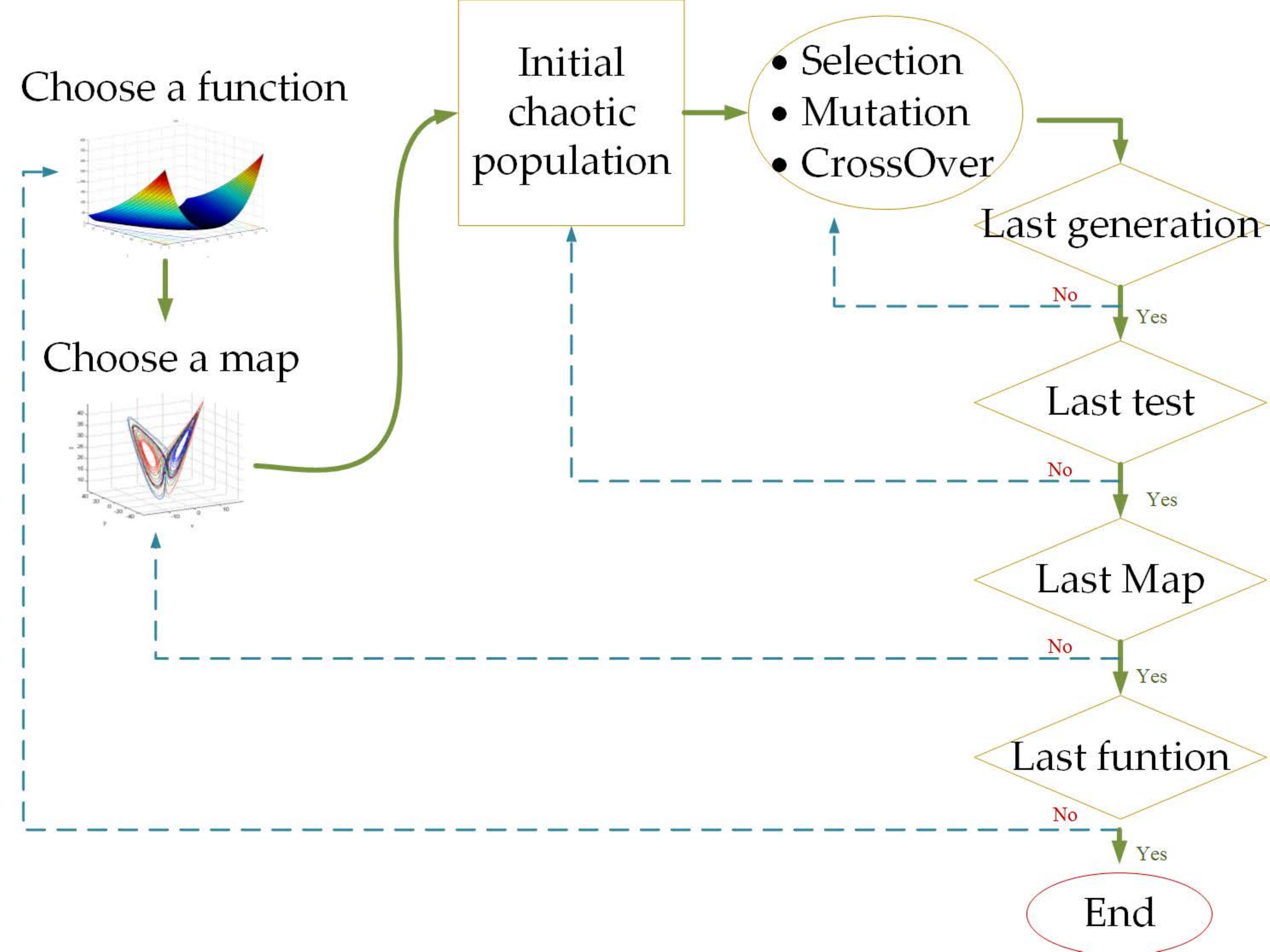}
	\caption{Diagram of the numerical experiment.}
	\label{fig:figura2}
\end{figure}

In Figure \ref{fig:figura2} we present the diagram of the algorithm, where a chaotic map is used for the generation of the initial population.

The implementation of the algorithm evaluates the ideal of each one of the nine benchmark functions, using eight different chaotic maps and a random function for the generation of the initial populations. We did 50,000 tests of every chaotic map by each function and then compare these results with the ones obtained by the traditional GA (that uses random series). The obtained results have been shown in a contour plot centered on the entropy of the initial populations.

\subsection{Initial Population}

The initial populations are generated with eight chaotic maps, \cite{Leontitsis2004} presented on the Table \ref{table:Parameters}, and a random normal function. For every routine of the algorithm, the entropy of the initial population and the fitness are calculated (see Table \ref{table:Algorithm}). 

\begin{table}[!h]
	\caption{Parameters of Chaotic Maps}
	\label{table:Parameters}
	\setlength{\tabcolsep}{3pt}
	\begin{tabular}{|p{55pt}|p{50pt}|p{50pt}|p{50pt}|}
		\hline
		Maps & 
		Parameter 1 & 
		Parameter 2 &
		Parameter 3 \\
		\hline
		Lorenz & $\alpha =10$ & $\varrho =28$ & $\beta=\frac{8}{3}$ \\
		Rossler & $\alpha =0.2$ & $b=0.2$ & $c=5.7$ \\
		Random & $\alpha =0$ & $b=1$ & \\
		Phaseran & $\alpha =1.95$ &  &  \\
		Mackeyglass & $\alpha =0.01$ & $b=1$ & $c=0.9$ \\
		Ikeda & $u=0.9$ &  &  \\
		Henon & $\alpha =1.4$ & $b=0.3$ &  \\
		Quadratic & $\alpha =1.75$ &  &  \\
		Logistic & $\alpha =4$ &  &  \\
		\hline

	\end{tabular}
\end{table}

\subsection{Flow of work}
The workflow for the search of the global optimum (for every evaluated function) is described in the following algorithm: 
{\color{blue}}

\begin{table}[h!]
	\caption{Chaotic Genetic Algorithm}
	\label{table:Algorithm}
	\setlength{\tabcolsep}{3pt}
	\begin{tabular}{l}
		\toprule
		 {\color{blue} \textbf {\footnotesize {ALGORITHM 1}}}: Structure of the proposed \\
		 ~~~~~~~~~~~~~~~~~~~~~~~~ Chaotic Genetic Algorithm.	\\
		 \midrule
		1~~Map for the generation of the initial population	\\
		2~~Initiate a set of tests for these maps	\\
		3~~Generate the initial population	\\
		4~~Calculate the entropy to the initial population	\\
		5~~Calculate the fitness.	\\
		6~~Save the greatest fitness.	\\
		7~~Produce.	\\
		~~~~7.1~~Process of selection through the roulette wheel.	\\
		~~~~7.2~~Mutation	\\
		~~~~7.3~~Crossing	\\
		8~~Go back to step 5, Gen = Gen + 1	\\
		9~~Go back to step 3, Test = Test + 1	\\
		10~Go back to stop 1, Map = Map + 1	\\
		11~End	\\
		\bottomrule	
	\end{tabular}
\end{table}

\section{Results}
The results obtained were grouped by each chaotic map in pairs (fitness and entropy). Figure \ref{fig:figura3} shows the distributions of the density of the results grouped into three categories:

\begin{figure*}[!t]
	\centering
	\includegraphics[width=0.80\linewidth]{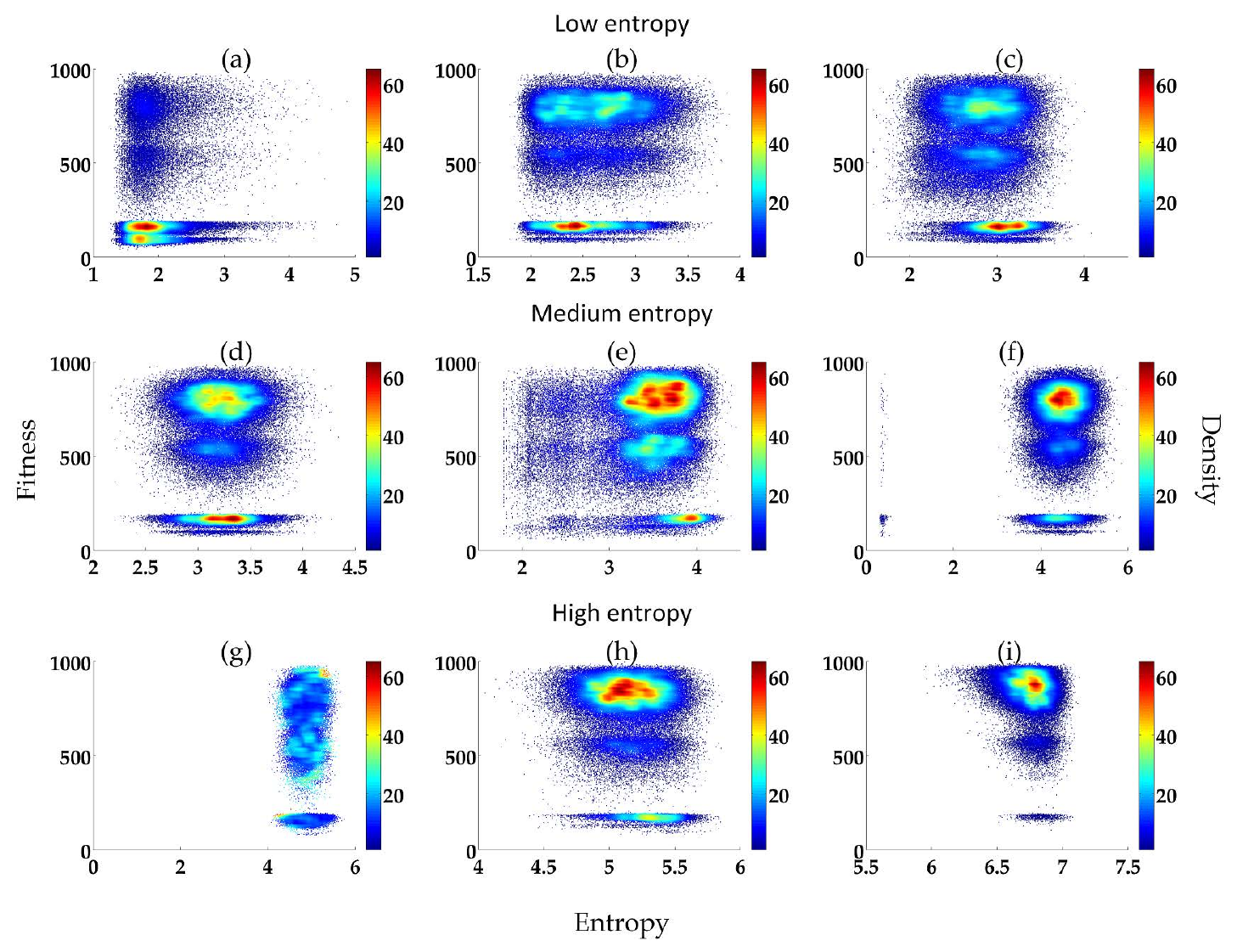}
	\caption{Density graphs for the different chaotic maps as a function of the entropy and fitness. The subfigures correspond to different maps: (a) Lorenz, (b) Rossler, (c) random function, (d)  Phaseram, (e) Mackeyglass, (f) Ikea, (g) Henon, (h) Quadratic and (i) Logistic map}
	\label{fig:figura3}
\end{figure*}

\begin{itemize}
	\item The map (c) corresponds to the distribution of the density of results using a random function for the generation of initial population. The Lorenz (a) and Rossler (b) maps display similar behaviors to map (c) regarding the fitness of the solutions. For this maps, the range of the entropy is concentrated between 1.5 and 3.0. 
	\item Secondly, we can observe a set of maps Phaseram (d), Mackeyglass (e), Ikea (f), Henon (g) and Quadratic (h) with distributions of densities of results of major fitness and with a range of the entropy of the maps of 3.0 $-$ 5.5.
	\item Thirdly, the Logistic map (i) obtained a distribution of densities of fitness with top values. The entropy of the initial populations has ranges superior to the rest of the maps 6.0 $-$ 7.0.	
\end{itemize}

\subsection{Analysis of the results}
The processing time study of every pair (benchmark function, chaotic map) and their execution times without significant variation of the average allows to compare the results obtained, through the analysis of the performances. Table \ref{table:Performance} presents the performance rates and  average entropies for all the benchmark function/chaotic map pairs. The performance corresponds to the percentage of the satisfactory results with an error rate lower than $\alpha$. The results present an increase of the algorithm's performance related to an increase on the average entropy of the chaotic map.

\begin{turnpage}
 \begin{table*}[!h]
	\caption{Parameters of Chaotic Maps}
	\label{table:Performance}
	\setlength{\tabcolsep}{3.45pt}
	\begin{tabular}{|l|c|r|c|r|c|r|c|r|c|r|c|r|c|r|c|r|c|}
		\hline
		\multirow{2}{*}{Chaotics Maps} & Overall  & \multicolumn{2}{|c|}{Ackley} & \multicolumn{2}{|c|}{Beale} & \multicolumn{2}{|c|}{Bukin6} & \multicolumn{2}{|c|}{Leon} & \multicolumn{2}{|c|}{Levil3} & \multicolumn{2}{|c|}{Matyas} & \multicolumn{2}{|c|}{Modschaffer2} & \multicolumn{2}{|c|}{Rastrigin} \\
		\cline{3-18}
		& performance & P & E & P & E & P & E & P & E & P & E & P & E & P & E & P & E \\
		\hline
Lorenz 		& 73.50 & 94 & 2.54386 & 0 & 0.195310 & 66 & 1.97144 & 98 & 1.88867 & 98  & 1.85878 & 98  & 1.97959 & 100 & 183.614 & 34  & 1.89743 \\
Rossler 	& 91.25 & 98 & 2.54386 & 82 & 2.67537 & 74 & 2.55464 & 92 & 2.61919 & 98  & 2.52400 & 100 & 2.63700 & 100 & 2.61023 & 86  & 2.62492 \\
Random 		& 89.00 & 94 & 2.83871 & 88 & 2.81170 & 64 & 2.88198 & 94 & 2.87322 & 98  & 2.85132 & 100 & 2.85118 & 88  & 2.87730 & 86  & 2.82495 \\
Phaseram 	& 91.50 & 90 & 3.15452 & 96 & 3.16689 & 68 & 3.21175 & 96 & 3.21292 & 100 & 3.22044 & 100 & 3.14271 & 100 & 3.22925 & 82  & 3.20808 \\
Mackeyglass & 95.75 & 92 & 3.19086 & 98 & 3.14796 & 88 & 3.15724 & 94 & 3.13313 & 94  & 3.36957 & 100 & 3.16983 & 100 & 3.26629 & 100 & 3.01408 \\
Ikeda 		& 93.75 & 98 & 4.44601 & 96 & 4.56970 & 72 & 4.57927 & 98 & 4.48140 & 100 & 4.48592 & 100 & 4.51327 & 100 & 4.41893 & 86  & 4.54659 \\
Henon 		& 93.75 & 96 & 4.85867 & 96 & 4.87027 & 90 & 4.88434 & 96 & 4.82141 & 100 & 4.84103 & 100 & 4.90713 & 100 & 4.86108 & 72  & 4.85701 \\
Quadratic 	& 92.25 & 96 & 2.86558 & 96 & 2.85629 & 72 & 2.69795 & 94 & 2.81000 & 98  & 2.86922 & 100 & 2.89237 & 100 & 2.83040 & 82  & 2.86725 \\
Logistic 	& 95.50 & 98 & 6.75772 & 92 & 6.75160 & 90 & 6.71270 & 90 & 6.76206 & 100 & 6.76206 & 98  & 6.75195 & 100 & 6.77400 & 96  & 6.76066 \\
\hline
\multicolumn{18}{l}{P: performance. E: entropy. $\alpha$: 0.001}
			\end{tabular}
\end{table*}
\end{turnpage}

% The stochastic nature of the GA, modified by initial populations of characteristics deterministic features, achieved increasing performances for all the functions in comparison the stochastic algorithm.

The use of chaotic maps for the generation of the initial populations in the genetic algorithm, increases considerably the performance of the algorithm in comparison to the tradition stochastic algorithm.

In addition, the proposed algorithm was also modified using chaotic maps in the mutation and the population processes. However, the results obtained using these modifications do not show significant improvement.

In Table \ref{table:Performance} the rows are the chaotic maps used in the numerical experiments, where they have been sorted in order of increasing average entropy. The columns are the different benchmark functions  and their corresponding performance rates and average entropies.

Finally, the analysis of the results of fitness through a contour plot presents a similar behavior of the initial populations indistinctly of the chaotic map. To greater entropy of the initial population better is the density of ideal solutions. The increase of entropy of the initial populations contributes a major level of information for the obtaining of a major quantity of optimal fitness.

\begin{figure}[!t]
	\centering
	\includegraphics[width=0.9\linewidth]{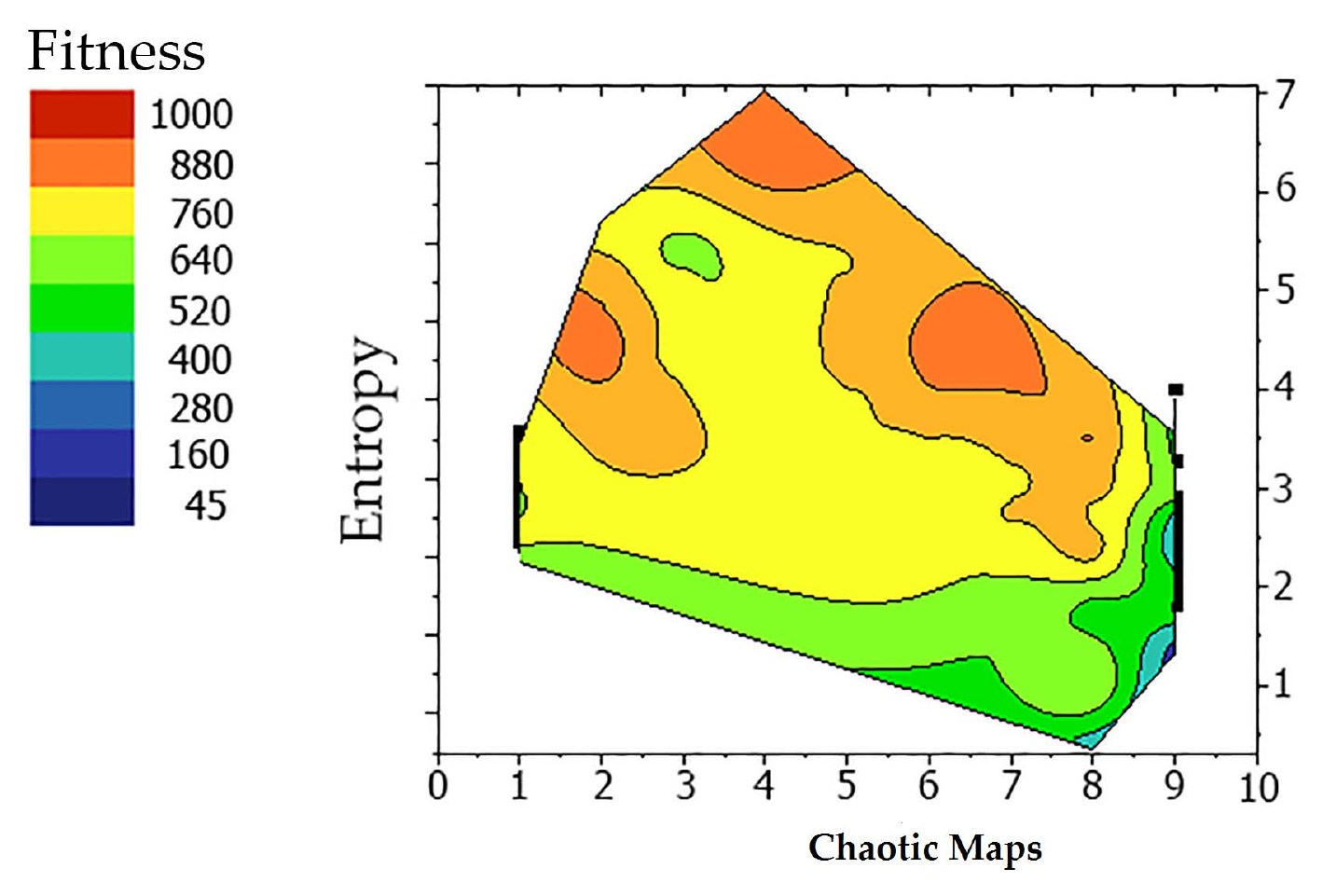}
	\caption{Contour plot of entropy versus chaotic maps. The order of the chaotic maps is Random (1). Quadratic (2). Henon (3). Logistic (4). Ikeda (5). Phaseram (6). Rossel (7). Mackeyglass (8) and Lorentz (9).}
	\label{fig:figura4}
\end{figure}

The contour plot of Figure \ref{fig:figura4}, it is organized with the chaotic maps of minor entropy in the ends of the graph for better visualization of the phenomenon. The order of the chaotic maps is Random (1). Quadratic (2). Henon (3). Logistic (4). Ikeda (5). Phaseram (6). Rossel (7). Mackeyglass (8) and Lorentz (9).

All the presented analyses were realized into 50000 tests by the couple (function. map) having 120 hours of processing with MatLab (Natick, Massachusetts, United States).

\section{Conclusion}

In this work, we study the Shannon entropy of the initial population used in the CGA. We found a strong relationship between the entropy of the initial populations and the densities of fitness of the solutions.

In general, all the chaotic maps, excepting of the curve of Lorenz's map showed the same behavior; the chaotic maps with higher entropies show an increase in the fitness's densities in the areas with better solutions: high entropies generated better solutions.

To check the validity and performance of the proposed approach, four experimental (numerical) studies were realized by sets of 50,000 routines for each of the eight chaotic maps. For this task, we used the MATLAB software in a cluster of 25 computers. Each experimental study is defined by a test set of nine mathematical functions.

The results presented here showed that the CGA is efficient solving complex problems functions from continuous nonlinear convex functions. As a future work, we plan to study the evolution of the average entropy and the influence of multifractals in the formation of the initial populations.

\section*{Acknowledgment}

This was supported in part by DICYT (Scientific and Technological Research Bureau) and the Department of Industrial Engineering of The University of Santiago of Chile (USACH). RS-G thanks FONDECYT (Chile) grant No.11160542 for financial support.

%\bibliographystyle{ieeetr} % estilo de la bibliografía. acm, apalike, ieeetr, plain
%\bibliography{library}

\end{document}